# UTILIZING ARTIFICIAL NEURAL NETWORKS TO PREDICT DEMAND FOR WEATHER-SENSITIVE PRODUCTS AT RETAIL STORES

**Elham Taghizadeh**
**Ph.D. student, Industrial Engineering Department, Wayne State University, U.S.**
Email: elham.taghizadeh@wayme.edu

___________________________________________________________________________________

**Abstract**
One key requirement for effective supply chain management is the quality of its inventory management. Various inventory management methods are typically employed for different types of products based on their demand patterns, product attributes, and supply network. In this paper, our goal is to develop robust demand prediction methods for weather sensitive products at retail stores. We employ historical datasets from Walmart, whose customers and markets are often exposed to extreme weather events which can have a huge impact on sales regarding the affected stores and products. We want to accurately predict the sales of 111 potentially weather-sensitive products around the time of major weather events at 45 of Walmart's retail locations in the U.S. Intuitively, we may expect an uptick in the sales of umbrellas before a big thunderstorm, but it is difficult for replenishment managers to predict the level of inventory needed to avoid being out-of-stock or overstock during and after that storm. While they rely on a variety of vendor tools to predict sales around extreme weather events, they mostly employ a time-consuming process that lacks a systematic measure of effectiveness.
We employ all the methods critical to any analytics project and start with data exploration. Critical features are extracted from the raw historical dataset for demand forecasting accuracy and robustness. In particular, we employ Artificial Neural Network for forecasting demand for each product sold around the time of major weather events. Finally, we evaluate our model to evaluate their accuracy and robustness.

## Introduction
Over the past few years Big Data has become the focus of IT innovation in business, science and industry. In general, Big Data means having the ability to gather, store, explore, and analyze with velocity, variety, and volume of data. Analytics refers to the ability to explore insight data by considering many known methods and techniques, such as statistics, mathematics, econometrics, simulations, and optimizations, to help business organizations and companies become lean and make a decision based on quick changes in the world. Most companies, industries, and organizations want to increase their ability to process their data (internal or external) efficiently. Some lean companies such as Walmart, eBay, Progressive Insurance, and Target have reported the ability to consider Big Data and its benefits; they store internal data such as transactions, last year's sales, etc. in their databases automatically (Wang et al, 2016).These companies are successful in extracting new insights and obtaining new forms of value by considering Big Data analysis in their operation plans (Sanders, 2016). This literature review discusses previous research on routing and scheduling logistics as well as the optimization operational network used in inventory and labor scheduling. Recently, other Big Data analytics applications, such as segmenting suppliers, measuring and mitigating risk, forecasting demand and informing suppliers have increased especially in the area of in sourcing (Kourentzes et al, 2014). Since a key aspect of supply chain management is designing and controlling processes, operations and inventory to meet demand, then forecasting demand by considering Big Data analysis is a huge step to deal with variations to decrease shortage or overproduce.

By increasing the application of Big Data among many companies, choosing the best tools and methods to analyze them is the key issue to consider. Because of the growth of computing power and the accessibility of data, the use of Artificial Neural Networks (ANN) for forecasting concepts is now commonly applied (Kourentzes et al, 2014). ANN is a mathematical model that includes the group of connected artificial neurons, which is like the human neural network, to compute and store specific information. This method has the ability to investigate and submit the nonlinear relationships of the given information and data, so ANNs have been widely considered as one  method or technique for pattern recognition, image processing, data mining, time series forecasting and so on (Du et al, 2014; Joghataie and Dizaji, 2016, 2013; Farrokh, M. et.al, 2015). ANNs enhance their forecasting accuracy and robustness by using these features as ensembles of several network models to cope with sampling and modeling (Kourentzes et al, 2014). This paper proposes an ANN to forecast the demand for weather-sensitive products based on external data (weather and date) and show its power to reach accrued forecasting.

With an increase in technology, Big Data can improve supply chain collaboration and operations such as inventory management, transportation (routing) management, and relationship management (Waller and Fawcett 2013). Sadovskyi et al. (2014) show that a significant lack of new theoretical and practical developments exist in this concept. Big Data can be used for the forecasting of delivery times and route optimization by looking to various types of data such as traffic and weather information. The analysis of sensor information for inventory and merchandising decisions is the other significant contribution of Big Data applications (Waller and Fawcett 2013). Lim et al (2014) shows that the most frequent application of Big Data in Supply Chain (SC) includes general subjects such as why use Big Data, how to use Big Data etc. The second applications are transportation (freight and public transport) and sales. Despite these uses of Big Data, it has not been used for improving the forecasting demand process. In this paper, we will consider different data such as weather to forecast customer demand for weather-sensitive products.

Use of predictive analytics and time-series methods is required for demand forecasting on independent products (Cheikhrouhou et al., 2011; Li et al., 2012). Exponential smoothing as a time-series method is widely considered for both short-term, intermediate-range and long-term forecasting because it can easily deal with both trends and seasonality. The auto-regressive model, which obtains a forecast in one period by calculating a weighted sum of the demand in previous periods, is another popular method for all types of forecasting. Moreover, associative forecasting methods are used for intermediate-range forecasting in service industries or manufacturing (Lu and Wang, 2010; Beutel and Minner, 2012; Taghizadeh et al., 2012). Optimizing techniques of demand or sales forecasting and operations planning provide an aggregative and fundamental management capability for marketing, production, sourcing, and especially inventory management to ensure customer satisfaction (Chen et al., 2010; Sodhi and Tang, 2011; Lim et al., 2014; Taghizadeh and Setak, 2013). Over the last two decades, the use of NNs for demand forecasting has increased (Zhang et al., 1998). ANNs have been proven to be general approximates and flexible for nonlinear data and prediction (Hornik, Stinchcombe, & White, 1989; Hornik, 1991). ANNs are able to fit any underlying data generating process, predict with high accuracy, and forecast both linear (Zhang, 2001) and nonlinear (Zhang, Patuwo, & Hu, 2001) time series of various forms. Researchers have tried to introduce several types of NNs and their applications over many years (for example, see Connor, Martin, & Atlas, 1994; Efendigil, Önüt, & Kahraman, 2009; and Khashei & Bijari, 2010). A variety of ANNs have been used in forecasting approaches: for instance, the single multiplicative neuron (SMN) model used for TSF (Yeh, 2013); the generalized regression neural network (GRNN) introduced as an effective and novel method in the time series forecasting (Yan, 2012); and radial basis function neural network (RBFNN) also used for TSF problems (Du & Zhang, 2008). Generally speaking, during the course of forecasting, minimizing errors is the important objective accomplished through different machine learning algorithms (Yeh, 2013). However, there is another requirement that is considered besides these errors; that is the number of hidden layer nodes. Du et al (2014) developed multiobjective evolutionary algorithms related to TSF problems. One of them, called nondominated sorting adaptive differential evolution (NSJADE), introduced knee point to time series forecasting problems for the first time. They showed that the Knee Point-NSJADE was more effective and competitive to predict time series databases. Kourentzes et al (2014) proved a new and comparative fundamental ensemble operator for neural networks that is based on estimating the mode of the forecast distribution. They showed that both median and mode are very useful operators. They used the mode and median as operators of ensemble research.

In this paper, we consider the external data which come from outside of the supply chain network and are gathered or stored by other organizations for demand forecasting. We show how these data can increase the accuracy of our demand forecasting. Since our data is huge and there is an unknown and nonlinear relationship between features and demands, we employ the single hidden layer Artificial Neural Network, for each product sold around the time of major weather events. Finally, we compare our results to show the impact of using external data in forecasting demand by applying ANNs. The organization of this paper is as follows. In Section 3, we review the feature selection method (Random Forest) to select important variables and features to improve our prediction. In Section 4, our mythologies for forecasting are introduce to predict sold units in different periods of time. In Section 5, we explore our data and pre-process the dataset. In Section 6, we discuss our results. Finally, conclusions are given in Section 7.

**Feature Selection Method**

Recently the size of datasets used in machine learning have increased rapidly. Because of the growth of information storage, data have a huge number of features, which can be irrelevant or redundant (Liu, J. et al, 2017). Therefore, feature selection methods can eliminate this information and play a very important role in handling large datasets (Seijo-Pardo et al., 2017). In addition, by applying an optimal evaluation criterion we can reduce the dimension of features compared to the number of original feature numbers, but in the selected feature subset we will obtain as much information as possible (Almuallim.H, & Dietterich .T, 1991; Hoque .N et al, 2014). Feature selection methods can be divided into three models based on the evaluation criteria (Chen .G, & Chen .J, 2015): the wrapper model, the embedded model, and the filter model. In the wrapper model, the performance of the feature selection depends on the classifier directly, such as IWSS (Bermejo .P et al, 2014) GA/FDA (Chiang .L, & Pell .R, 2004) and RFE (Guyon .I, Weston .J, Barnhill .S, et al., 2002).For the embedded model, we need a specific learning algorithm before conducting the feature selection in our process which can be viewed as in an intermediate method between wrappers and filters. The filter model is used to maximize the evaluation function for getting an optimal feature subset through a search strategy (Liu, J. et al, 2017).

One of the main issues in applying the feature selection is considering an appropriate method for specific problems. Each feature selection method has advantages and disadvantages, and its performance depends on the dataset. However, despite the availability of the growing number of methods, generally no specific feature selection method exists. Some knowledge of existing algorithms is generally required in order to be able to choose a method that is appropriate to the problem. One possible solution to this problem is to use an ensemble method since better results could be achieved by combining different machine learning methods to solve the same problem; thus, We used a known ensemble model, Random Forest (RF) (Seijo-Pardo et al., 2017). Random Forest (RF) is a kind of supervised classifier constructed from an ensemble of classification and regression trees (CART) that use most of their constituent terminal nodes to forecast the class of a given data(Seijo-Pardo, B. et al, 2017).

## Methodology

**Artificial Neural Networks (ANN)**
ANNs are information processing systems that simulate the behavior of the human brain (Martín del Bío & Sanz Molina, 2006). ANNs obtain the inherent information from the considered features and learn from the input data, even when our model has noise (Kasabov, 1996). ANN structure is composed of essential information processing units, which are neurons. They are defined into several layers and interconnected with each other by defining weights. Synaptic weights show the interaction between every pair of neurons (Garrido, C. et al, 2014). These structures distribute information through the neurons. The mappings of inputs and estimated output responses are calculated through combinations of different transfer functions. We can use the self-adaptive information pattern recognition methodology to analyze the training algorithms of the artificial neural networks. The most commonly used computation algorithm is the error back propagation algorithm proposed by the PDP group of Rumelhart in 1985 (Lippmann, R. P, 1987; Sharma, A., & Panigrahi, P.K, 2011).

Neural networks can be divided into single-layer perception and multilayer perception (MLP) networks. The multilayer perception network includes multiple layers of simple, two state, sigmoid transfer functions having processing neurons that interact by applying weighted connections. A typical feed-forward multilayer perception neural network consists of the input layer, the output layer, and the hidden layer as shown in Exhibit 1 (Sharma, A., & Panigrahi, P.K, 2011). The multilayer perceptron (MLP) with the back propagation learning algorithm is used in this study because numerous previous researchers used this type of ANN (Gedeon, Wong, & Harris, 1995), and it is also a general function approximator (Funahashi, 1989).

**Exhibit 1.** Architectural Graph of an MLP Network with Two Hidden Layers.

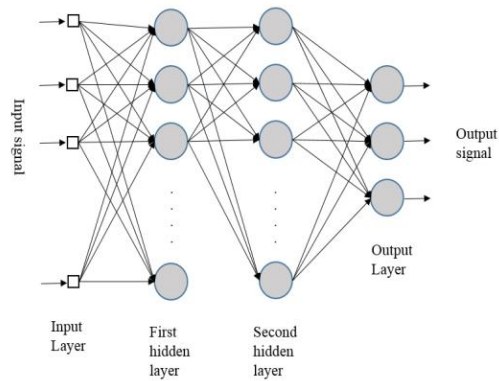

**Time-Delay Neural Networks**
The time-delay structure (Lollia, F. et al, 2017) differs from the feedforward one because each input t not only consists the feedforward network input $y_t$ but also some of the preceding samples. The number of these preceding inputs is called taps (Lollia, F. et al, 2017). Therefore, the time-delay neural network input t is shown in Equation (1):

$$\hat{y_t} = (y_t, y_{t-1}, \ldots\ldots, y_{t-taps}) \quad (1)$$

This changes both the dimension of each input vector and of each input weight vector ($R^{n*(taps+1)}$). The rest of the architecture remains the same.

**Recurrent Neural Networks (RNN)**
The Recurrent structure applied in this study forecasting as shown in Exhibit 2 (Lollia, F. et al, 2017) includes a layer connected to the hidden layers and empowers the training step using the pervious output signals. In this structurer, each input t includes both the sample $x_t$ and hidden layer's output defined as:

$$\widehat{g_t} = (g(w_1.\hat{x}_t + b_i), g(w_2.\hat{x}_t + b_i), \ldots\ldots g(w_N.\hat{x}_t + b_i)) \quad (2)$$

The Recurrent architecture was introduced by Elman and Zipser (1988) to add dynamic memory to our network. Nasiri Pour et al. (2008) tested it with promising results for forecasting demand pattern.

**Exhibit 2**. Architectural Graph of Recurrent Neural Networks (RNN).

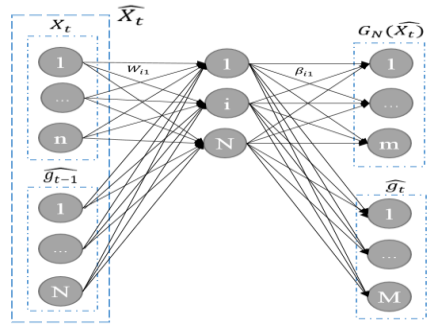

**Bagging**
When small changes in learning data occur, the decision trees become unstable. If the first cutting variables are different because of a minor change in the learning data, the entire structure of the tree may be modified. Bagging, as well as a tree-based ensemble methods, employs the fact that singular trees may produce unstable results but the correct prediction on average. Bagging trains a number of trees on a boot and applies all the constructed trees on the test set. The final prediction is the average value of the predictions resulting from each tree (Wauters, M., & Vanhoucke, M., 2016). The superiority of bagging over singular classification or regression trees was introduced by Bühlmann and Yu (2002). In that paper, the authors analyzed the reduction of variance by using bagging. For forecasting demand, We used bagging as an ensemble method in this study and compared its results with other methods.

## Experimental Analysis

**Data Collection**
To empirically evaluate the performance of the methods described in pervious section, a daily data set collected at 45 different Walmart stores from 2012 to 2014 is used (the original data set is available at https://www.kaggle.com/c/walmart-recruiting-sales-in-stormy-weather/data ) . In this paper, We want to accurately predict the sales of 111 potentially weather-sensitive products (such as umbrellas, bread, and milk) around the time of major weather events at 45 of their retail locations. We used the information from 20 stations where weather features were calculated from 1/1/2012 until 10/31/2014. In applying the ANNs already emphasized in pervious section, the data set was divided into a training and a test set. the training data included the history of unit sales for 111 potentially weather-sensitive products from 1/1/2012 to 05/31/2014 (contained in 3892716 observations), and the training data set included the rest of them from 06/1/2014 to 10/31/2014 (around in 721000 observations). Because the amount of unit sales for all 111 different products is wide ranging and highly fluctuating, the best method to evaluate our prediction models will be the Root Mean Squared Logarithmic Error (RMSLE), Equation (3) is shown below.

$$\sqrt{\frac{1}{n}\sum_{i=1}^{n}(\log(p_i + 1) - \log(a_i + 1))^2} \tag{3}$$

To follow the Equation (3) We changed my unit sales (considered as output for my model described in section 4) to log unit sale and replaced them with the amount of unit sales in my data set. Exhibit 3 plots log unit sale for all 111 different types of products during our time line, and is obvious that some products have the same behavior during these times, so we can conclude that they may belong to the same categories; for example, products 1, 2, 3, and 4 may

**Exhibit 3.** Daily Scatter Plot Log Unit Sale.

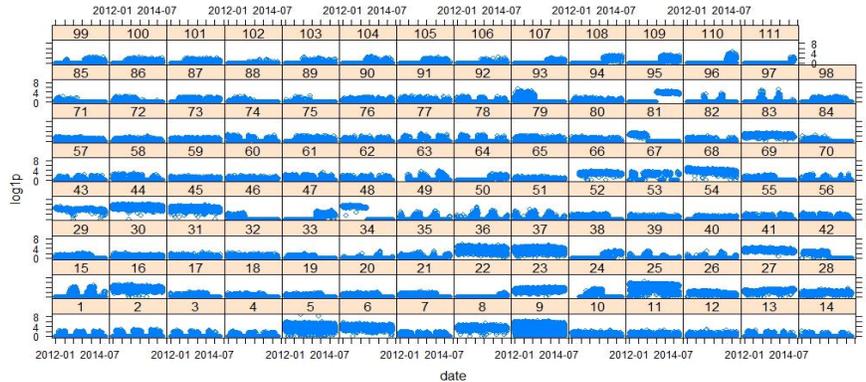

be sensitive to a specific season (winter or summer). In the next step for easily predicting and removing non-informative information, We excluded products at all stores whose log unit sales are zero overtime since customers might not tend to purchase these products on specific dates. Also We removed all data or information on the date 2013-12-25 since it was a holiday and all Walmart stores were closed. Therefore, my observation's testing dataset is reduced to 235789. Finally, in my testing data, their log unit sale (unit sale) would be predicted to be zero; then, our model just focused on predicting the remaining $\log unit\ sale$. After reducing my data set to informative data by removing the zero units, We defined two data sets as follows to compare whether using ANNs and a weather feature in forecasting will help us have more accurate predictions:
1. Data set 1 included input date and weather features, output log unit sale Add (specified by Ddate Dweather)
2. By considering that events and holidays could affect customer demands, data set 2 included input date, weather features, and holiday (or event) features obtained as in the following steps. (specified by Devent)
    2-1. weekday, is_weekend, is_holiday, is_holiday_and_weekday, is_holiday_and_weekend
    2-2. is_BlackFriday-3days, -2days, -1day, is_BlackFriday, +1day, +2days, +3days
    2-3. year, month, day

**Random Forest**
There were some unrelated or unimportant variables in our data sets 1 and 2; thus, as described in pervious sections, we applied Random Forest (4 models: ntry 2and 4, ntree 50 and 100) to define and identify the essential features. Exihibit 4 shows the results (MSE shows mimum square error of respective model and % Var explained(%IncMSE) are outcome of applying Random Forest) and we concluded that the Random Forest with ntry= 4 and ntree=50 has minimum error, so we chose it as our model for feature selection. Our Dweather dataset has 20 variables and Devent dataset has 31 variables (included 20 variables of Dweather and events which are described in pervious section). By comparing the value of %IncMSE (is the importance measures which shows how much MSE or Impurity will increase when that variable is randomly removed) which is one of outcome of our model, we chose and kept variables with the absolute value %IncMSE more than 1. Finally, for the Dweather dataset the number of variables decreased to 15 and for the Devent the number of variables decreased to 22. By reaching the optimized features for all data sets, we fitted our models to log *unit sale* values by using methodologies described in pervious section.

**Exhibit 4.** Random Forest Results.

| Model | MSE | | | | % Var explained | | | |
|---|---|---|---|---|---|---|---|---|
| | ntry:2 ntree:50 | ntry:2 ntree:100 | ntry:4 ntree:50 | ntry:4 ntree:100 | ntry:2 ntree:50 | ntry:2 ntree:100 | ntry:4 ntree:50 | ntry:4 ntree:100 |
| **Dweather** | 2.83338 | 2.876044 | 2.65173 | 2.700654 | 8.34 | 6.96 | 14.22 | 12.63 |
| **Devent** | 1.40582 | 1.130887 | 0.22281 | 0.223172 | 54.52 | 63.42 | 92.79 | 92.78 |

After applying random forest to keep imformative variables, We fitted different ANN models to forecast the log (unit sales) by applying bagging, timing lag, RNN and MLP. For constructing the MLPs, we defined layers ranging 2 to 4 with the number of neurons ranging from 20 to 100 that present in Exhibit 5 (column Methodology). We summarized the MLP structure by using three notations; for example, MLP-L2-N30 means this MLP has 2 layers and 30 neurons at each layer. In Exihibit 5, we compared the results of two datasets in separte columns (Devent and Dweather), for each of them we calculated MSE and time (two important factors in prediction methods) related to specific ANNs models which were categorized in first column (Methodology). In the following section we discussed in details how we chose an appropriate model and predicted the number of unit sales.

**Exhibit 5.** MSE Through Different MLP Models.

| | Devent | | Dweather | | | Devent | | Dweather | |
|---|---|---|---|---|---|---|---|---|---|
| **Methodology** | MSE | Time (mint) | MSE | Time (mint) | **Methodology** | MSE | Time | MSE | Time |
| MLP-L2-N20 | 0.324734 | 2.6535 | 3.482135 | 2.33 | MLP-L3-N70 | 0.200235 | 7.303 | 2.195085 | 6.4 |
| MLP-L2-N30 | 0.309068 | 3.2508 | 3.314149 | 2.85 | MLP-L3-N80 | 0.2081061 | 9.9939 | 2.095009 | 8.7581 |
| MLP-L2-N40 | 0.295530 | 4.0144 | 3.168985 | 3.52 | MLP-L3-N90 | 0.21464939 | 12.3273 | 2.07501 | 10.8029 |
| MLP-L2-N50 | 0.28299 | 4.3370 | 3.034576 | 3.8 | MLP-L3-N100 | 0.21681627 | 19.1324 | 2.098464 | 16.7665 |
| MLP-L2-N60 | 0.271233 | 5.0735 | 2.731941 | 4.45 | MLP-L4-N20 | 0.3109816111 | 3.2583 | 3.07141 | 2.8198 |
| MLP-L2-N70 | 0.268715 | 6.4514 | 2.881444 | 5.65 | MLP-L4-N30 | 0.2959791966 | 3.9918 | 2.923239 | 3.4546 |
| MLP-L2-N80 | 0.255970 | 8.8286 | 2.74478 | 7.74 | MLP-L4-N40 | 0.2830148888 | 4.9295 | 2.795197 | 4.2662 |
| MLP-L2-N90 | 0.264018 | 10.8899 | 2.831082 | 9.54 | MLP-L4-N50 | 0.2710111528 | 5.3256 | 2.676642 | 4.6089 |
| MLP-L2-N100 | 0.266684 | 16.9014 | 2.666583 | 14.8 | MLP-L4-N60 | 0.2597462461 | 6.23 | 2.565384 | 5.3916 |
| MLP-L3-N20 | 0.241977 | 2.7145 | 2.594735 | 2.38 | MLP-L4-N70 | 0.2573352741 | 8.766 | 2.541572 | 7.5864 |
| MLP-L3-N30 | 0.230304 | 3.3255 | 2.46956 | 2.91 | MLP-L4-N80 | 0.2451301522 | 11.9961 | 2.421028 | 10.3818 |
| MLP-L3-N40 | 0.220216 | 4.1068 | 2.29808 | 3.6 | MLP-L4-N90 | 0.2528375557 | 14.7970 | 2.49715 | 12.8057 |
| MLP-L3-N50 | 0.210876 | 4.4367 | 2.297093 | 3.89 | MLP-L4-N100 | 0.2553899442 | 22.9654 | 2.522359 | 19.8749 |
| MLP-L3-N60 | 0.202111 | 5.1902 | 2.220553 | 4.55 | | | | | |

## Discussion

For each defined MLP in Exihibit 5, our inputs were our variables and output was log *unit sale* values. We ran each model 25 times and calculated the average MSE (Mean Square Errors); Exhibit 5 summarizes the results. By looking through Exihibit 5 we can found that by increasing the number of layers, the MSE started to decrease, but after changing 3 layers to 4 layers, the MSE started to increase. It was proved that by increasing the number of layers after 3 layers, we did not have improvement in our model (it means we could not have lower MSE).

You can see the minimum error belongs to MLP-L3-N70 for Devent and MLP-L3-N90 for Dweather, so we can use these MLP structures to forecast our unit sales. We fixed these models (MLP- L3-N70 for Devent and MLP-L3-N90 for Dweather) and compared their training results with others (RNN, Bagging and time lagging) to find an appropriate model.

We ran RNN and time lag model with same structure that we fixed for MLP for each Data set (3 layers and 70 Neurons for Devent and 3 Layers and 90 Neurons for Dweather). Exihibit 6 shows the results of training for RNN and timing lag. Also, we ran bagging for 10 times and saved the average of its training results in Exihibit 6 (in Exihibit 6 different type of ANN models were categorized in the first column and we compaired the MSE and time of training of each data set in diffirent columns and tables related to each models) . By comparing training MSEs in Exihibit 6, the appropriate model with minim MSE is belong to MLP for both dataset. However, when we compare the computational time, we can find that acceptable error with a minimum or appropriate time is belong to the bagging. Generally, the main goal in most industries is to reach minimum error in forecasting, so we chose and applied MLP to predict two data sets (Dweather and Devent), and it is obvious their times are not so affective. In addition, Exihibit 6 presents how the timing lag and RNN with maximum errors are absolutely not an acceptable method for forecasting compared to others since our output may not be influenced by previous information and just tends to be related to features and variables.

Finally, we tested MLP (MLP-L3-N70 Devent, MLP- L3-N90 Dweather) and other methods by using the testing data to make sure to choose the appropriate model for forecasting, and their results summarizes in Exihibit 6 in Test columns related to each model. The MSE of the testing data set was obtained as follows: Devent (0.198765) and Dweather (2.000165) and confirm our accuracy to use these models for forecasting the Devent and Dweather data set. From Exihibit 6, it appears that by considering event features in addition to weather features, we have more accurate forecasts with smaller errors, but their differences are not too great. This may show that these products are affected by event more than weather features. We also used simple linear regressions for forecasting the $\log unit\ sale$, and the mean square error obtained was 12.88769, which is too high to compare with MLP. Therefore, it may show that by considering weather and event features in addition applying MLP, we could have a more accurate model with minimal error.

**Exhibit 6.** Compare Results of Different Model.

| Dweather | | | | |
|---|---|---|---|---|
| **Model** | **Train** | | **Test** | |
| | **MSE** | **Time** | **MSE** | **Time** |
| **MLP-L3-N90** | 2.07501 | 10.8029 | 2.000165 | 8.8764 |
| **Time-Delay** | 4.440019 | 16.0045 | 4.85769 | 14.7863 |
| **RNN** | 5.476349 | 20.8756 | 5.67898 | 18.9765 |
| **Bagging** | 2.550736 | 3.98761 | 2.764989 | 3.2145 |
| Devent | | | | |
| **Model** | **Train** | | **Test** | |
| | **MSE** | **Time** | **MSE** | **Time** |
| **MLP-L3-N70** | 0.200235 | 7.303 | 0.198765 | 6.9887 |
| **Time-Delay** | 0.486539816 | 19.0098 | 0.46780987 | 15.8703 |
| **RNN** | 0.61287609 | 21.9873 | 0.54678295 | 19.0024 |
| **Bagging** | 0.2323396 | 4.0176 | 0.2298746 | 3.7654 |

Finally, when we chose my appropriate methodology to fit a model, its output is $\log unit\ sale$ values. Therefore, for obtaining the unit sale of each items we must follow the below steps:
1. Predicted unit = exp(predicted_log1p*) – 1
2. Assigned zero value for: item/stores whose units are all zeros and on date 2013-12-25.

\* obtain from MLP models.

Applying weather features can help retail stores predict product inventory and reduce any lack of inventory problems in the future.

## Conclusions and Future Research

Accurate forecasts with minimal errors are essential for an efficient inventory control system. Inventory management of weather sensitive products is one example of this, whereby changing weather conditions increases or decreases product demands, so retailers and industries need to consider other features to obtain an accurate model. Another factor which can help retailers obtain an appropriate forecast model is choosing the best methodology. The multilayer perceptron (MLP) with the back propagation learning algorithm neural network represents a promising approach; its main strength is the ability to control non-linear functions without the requirement for distribution assumptions. However, few papers investigated the forecasting accuracy for weather sensitive products, and they have not been widely applied in real world problems due to the great computational effort for training the networks. Back-propagation learning has been typically fitted into this research field. In order to improve the understanding of these predictors (MLP with the back propagation learning algorithm) in the field of forecasting to avoid the lack of inventory, this study provides a comparison between standard methods. This study shows that back-propagation has better performance, and thus should be suggested to practitioners. Conversely, timing lag, Bagging, Recurrent Neural Network (RNN) and linear regression do not have perfect performance. Our research had some limitations; for example, our data summarized a two year period, and it is better to consider a bigger data set; also our data was not collected very carefully especially weather features which were recorded at different stations and have several missing data. The main advantages and disadvantage of applying this approach (ANN) can be summarized in the following points:

1. In our study, the high capability of forecasting and its accuracy that characterizes ANNs are met in the used database, and this accuracy is much higher and more appropriate than those obtained by applying some other methods, such as bagging, regression models, and timing lag in the field of inventory management at retail stores.

2. The main disadvantage of this approach is the computational time, since a large number of ANNs must be trained and tested, and the back propagation learning algorithm determining the connection weight must be calculated as many times as ANNs are involved in the selected architecture.

For our future work, we will test larger databases to examine the performance of ANNs. Meanwhile, more information of different databases will be explored during the course of forecasting, like the distribution of the ANNs with the worst forecast results. The other future research can be manipulating or merging weather features, since in this study we applied the exact value of features but by transferring or merging weather features maybe we can obtain better results. Finally, ANNs will be widely used for forecasting problems in real-world industries to see if they are also efficient. ANNs are proposed in this research because of their numerous advantages over more traditional parametric models, but also over other non-parametric models, such as decision trees. ANNs are a better fit for the phenomenon under this study.

## About the Author(s)

**Elham Taghizadeh** is currently a PhD student in the Department of Industrial and Systems Engineering at the Wayne State University, Detroit, Michigan, USA. She received her MSc and BS in Industrial Engineering from the, K.N. Toosi University of Technology, Iran. Her research interests mainly include supply chain, heuristic and meta heurastic optimisation and Big Data analysis.